\documentclass{article}
\usepackage{graphicx} 

\title{A Comprehensive Empirical Evaluation of Existing Word Embedding Approaches}
\author{
  Obaidullah Zaland\\
  \texttt{obaidullahzaland1997@gmail.com}\\
  South Asian University
  \and
  Muhammad Abulaish\\
    \texttt{abulaish@sau.int}\\
    South Asian University
  \and
    Mohd. , Fazil\\
  \texttt{mohdfazil.jmi@gmail.com}\\
South Asian University
}
\date{June 2020}

\DeclareUnicodeCharacter{0301}{\dash}
\DeclareRobustCommand\dash{%
  \unskip\nobreak\thinspace\textemdash\allowbreak\thinspace\ignorespaces}
\usepackage[labelformat=simple]{caption,subcaption}

\usepackage{flushend}
\usepackage{amsmath}
\usepackage{tabularx}
\usepackage{tabu}
\usepackage{multirow}
\usepackage{booktabs}
\usepackage{multirow}
\usepackage{graphicx}
\usepackage{flushend}
\usepackage{algorithmic}
\usepackage[lined,ruled,linesnumbered]{algorithm2e}
\usepackage{amssymb}
\usepackage{url}
\usepackage{booktabs}
\usepackage{amsfonts}
\usepackage{extarrows}
\usepackage{multirow}
\usepackage{hhline}
\usepackage{color}
\usepackage{wrapfig}
\usepackage{hyperref}
\usepackage{float}
\usepackage{array}
\usepackage{longtable}
\usepackage[justification=centering]{caption}
\usepackage{comment}
\usepackage{booktabs}
\usepackage{ragged2e}
\usepackage{varioref}
\usepackage{textcomp}
\usepackage{algorithmic}
\usepackage[lined,ruled,linesnumbered]{algorithm2e}
\usepackage{eqparbox}
\usepackage{makecell}
\usepackage{mathtools}
\usepackage{lscape}
\usepackage{dblfloatfix}
\usepackage{ wasysym }
\usepackage[backend=biber,
style=numeric,
citestyle=ieee]{biblatex}

\begin{document}


\begin{abstract}

Vector-based word representations help countless Natural Language Processing (NLP) tasks capture the language's semantic and syntactic regularities. In this paper, we present the characteristics of existing word embedding approaches and analyze them with regard to many classification tasks. We categorize the methods into two main groups - Traditional approaches mostly use matrix factorization to produce word representations, and they are not able to capture the semantic and syntactic regularities of the language very well. On the other hand, Neural-network-based approaches can capture sophisticated regularities of the language and preserve the word relationships in the generated word representations. We report experimental results on multiple classification tasks and highlight the scenarios where one approach performs better than the rest.  
\end{abstract}


\maketitle

\section{Introduction}
Dense real-valued word vector representations have been essential to NLP tasks, such as question answering \cite{liu2017stochastic, xiong2016dynamic}, semantic role labeling \cite{zhou2015end, he2017deep, foland2015dependency}, textual entailment \cite{chen2016enhanced} and machine translation \cite{zhang2017context}. These word representations, also called word embeddings, encode semantic and syntactic characteristics of the words, so these embeddings can act as input features to downstream tasks like sentiment analysis \cite{severyn2015twitter, sun2019utilizing}, rumor detection \cite{geng2019rumor}, and fake news detection \cite{long2017fake}.\\
Since the early days of natural language understanding, an exhaustive amount of research has been put into obtaining these word representations from a corpus of unlabeled text. Researchers have used statistical models \cite{landauer1998introduction, blei2003latent}, neural network language models \cite{bengio2003neural}, log-bilinear models \cite{mnih2013learning}, and context-based neural models \cite{peters2018deep, devlin2018bert}, among others, to construct these word representations. Nevertheless, the question of which approach to use in a specific scenario remains unanswered. 
To compare word embedding methods, we can use intrinsic or extrinsic evaluators\cite{bakarov2018survey}. Intrinsic evaluators measure the quality of word embeddings directly using semantic and syntactic relationships among them\cite{baroni2014don}, while extrinsic evaluators measure the quality of these word embeddings on downstream tasks using them as input features. 

Words with closer meaning to each other are called semantically related terms, and the embeddings produced for such words shall lie closer in the produced embedding space. For example, words like \textit{bicycle}, \textit{cycle}, and \textit{bike} are similar and semantically related to each other. Syntactically related words are those words which are bind by the syntax of a natural language such as English. For example, \textit{brief} and \textit{briefly} are syntactically related to each other, where \textit{briefly} is the adverb for the adjective \textit{brief}. Also, \textit{big} and \textit{bigger} are syntactically related, where the latter is the comparative form of the former.

Most studies investigate word embedding models concerning their intrinsic characteristics. Although the intrinsic evaluation of these models sheds light on the semantic and syntactic similarities among these words, they do not explicitly depict scenarios where one model is superior or inferior to others. Moreover, intrinsic evaluators require additional resources in the form of pre-defined queries for subjective tests\cite{wang2019evaluating}. These queries can be word couplets, such as adjectives and their comparative forms, or countries and their capitals, and are also called \textit{query inventories}\cite{schnabel2015evaluation}. 

Extrinsic evaluation of these methods, on the other hand,  maybe computationally expensive and time taking, but it provides much more insight into the quality of the word representations that these methods extract. To evaluate the actual quality of these methods, we should compare these approaches with regard to their real-world use cases. Small factors like the window size, data pre-processing, skewness of dataset, and dimension of the word embeddings, sometimes have an impact on the result of downstream tasks (e.g., classification) \cite{yang2018using}. 

However, researchers have investigated the relationship between intrinsic and extrinsic evaluators; they have not been comprehensive\cite{chiu2016intrinsic}. Besides, these comparative studies have not evaluated all the existing models, as most studies only focused on examining LSA, word2vec, and GloVe \cite{naili2017comparative,dhingra2017comparative,suleiman2018comparative}. Furthermore, as per our knowledge, there has been no extensive extrinsic evaluation that involves all the state-of-the-art models (GloVe, Word2vec, ELMo, BERT and fasttext). 

An excellent word representation method should be able to take care of some essential points.
\begin{itemize}
    \item The most frequent words such as \textit{the} or \textit{and} should not affect the quality of the word vectors.
    \item Rare words should have quality word representations.
    \item Multiple word embeddings for multiple word senses.
\end{itemize}
Apart from the above qualities, intrinsic to the word embeddings, the features extracted by these models should work well in different scenarios; for example, if we are dealing with sentiment analysis, the features should work fine with balanced and unbalanced data. In this paper, we use pre-trained vectors, trained using the dominant word representation algorithms for different scenarios of classification tasks. We also train word embeddings on these algorithms from scratch to measure the effect of pre-training with respect to each algorithm in various tasks. Besides, we study the impact of model parameters (window size and embedding dimensions) on the output of the downstream tasks. The effect of external parameters on the output of these models is also studied. These external parameters are the degree of formality of corpora used for pre-training, the amount of text in the corpus used for pre-training, and pre-processing. \\
We organize the rest of the paper in the following sections. Section 2 introduces existing word representation models, including traditional and neural models. We present the properties of word embedding models and compare the existing models according to these properties in section 3. The results have been reported in section 4. We have carried out exhaustive comparisons on both pre-trained and trained word embeddings. In the end, we conclude our findings in section 5. \\
\section{State of the Art Word Embedding Approaches}

Word embeddings act as a backbone for the downstream natural language processing tasks. Hence, numerous approaches have been proposed over the years to train these embeddings. For clarity, we classify these models into two categories: traditional and neural network models. 

\subsection{Traditional models}

Traditional models construct the word representations from the statistical information present in the corpus on the basis of the idea of distributional semantics. They use term frequencies, term-term co-occurrence frequencies \cite{Lund1996}, and term-document frequencies \cite{landauer1998introduction} as the basis for these vector representations. One Hot Encoding, the simplest model, uses a vector of size $|V|$, where $|V|$ is the size of the vocabulary to represent each word. The word vectors produced, have a value 1, at a specific position for that word and 0 everywhere else. For example, if $V = \{have, a, great, day\}$, then the word \textit{have} can be represented as $\{1,0,0,0\}$, the word \textit{a} can be represented as $\{0,1,0,0\}$, and so on.  

Latent Semantic Analysis (LSA) \cite{landauer1998introduction} is the most influential model in this category. LSA utilizes the statistical information present in the corpus to build a term-document frequency matrix $X$. LSA then uses Singular Value Decomposition (SVD) to find a low-rank approximation to the matrix to construct the word vector representations. SVD decomposes the co-occurrence matrix $X$ into three matrices, $V, V^T$, and $\Sigma$. 
\begin{equation}\label{svd}
    \begin{split}
        X = V\Sigma V^T 
    \end{split}
\end{equation}
While $\Sigma$ is a diagonal matrix comprising the singular values of the matrix, $V$ and $V^T$ are orthogonal matrices, comprising of left and right singular vectors. For obtaining a lower-rank approximation of rank $j$, for the matrix $X$, we select the $j$ largest singular values alongside the right and left singular vectors from $ V $ and $ V^T $ corresponding to those singular values. 
\begin{equation}\label{lower-rank-svd}
    \begin{split}
        X_j = V_j \Sigma_j V_j^T
    \end{split}
\end{equation}

On the other hand, Hyperspace Analogue to Language (HAL)\cite{Lund1996} uses term-term frequencies to construct the co-occurrence matrix $X$. HAL passes a "window" over the text, and words within the window are termed to co-occur with a strength inversely proportionate to the number of terms between them in the window. Words occurring to the right and left of a word are recorded separately. As a result, a matrix with $n$ rows and $2n$ columns, $n$ being the size of the vocabulary, is formed. As the vocabulary grows, the co-occurrence becomes enormous, and so does the word embedding size. Hence, columns with higher variance are selected out of $2n$ columns to reduce the dimensionality of word embeddings. Columns related to the most frequent words such as \textit{the} have higher frequencies while providing little information about the text. These high-frequency or \textit{high variance} columns contribute disproportionately to the distance between produced word representation vectors. 

Correlated Occurrence Analogue to Lexical Semantic (COALS)\cite{rohde2006improved} uses a normalization procedure to minimize the influence of most common words on the quality of produced word embeddings. COALS removes different columns for left and right contexts and adds a single column for each word, making the co-occurrence matrix symmetric. Moreover, COALS uses conditional rate, that is, whether word \textit{a} co-occurs more or less with word {b} than in general, as matrix entries instead of raw term-term co-occurrence count. Pearson's correlation coefficient can be used to calculate the conditional rate between word pairs. After formulating the co-occurrence matrix, negative entries are removed from the matrix, and the positive values are replaced by their square roots. The authors explain that the negatively correlated terms may not relate to each other semantically and haven't been used in the text corresponding to the same topic. They argue that removing the negatively correlated word columns results in less information loss than removing low-variance word frequency columns.

A variety of traditional models use the corpus statistics in different manners to form the co-occurrence matrix. Moreover, apart from SVD, other transformation methods such as Hellinger Principal Composition Analysis (HPCA) \cite{lebret-collobert-2014-word} have been used to learn these word representations. Although traditional models are easy to understand, they are computationally expensive and become infeasible, to work with on large datasets\cite{altszyler2016comparative}, because the co-occurrence matrix becomes huge and impractical to operate on. 

\subsection{Neural Models}

Unlike traditional models, neural models have evolved extensively with time. The first generation of these models was Neural Network Language Models (NNLM). Although initial Neural Network Language Models \cite{bengio2003neural} solved the curse of dimensionality problem present in statistical models, noticeable gains came after the introduction of Recurrent Neural Network Language Models (RNN-LM) \cite{mikolov2010recurrent, mikolov2011extensions}. In a natural language text, language modeling predicts the word sequence $w_1w_2...w_T$ probability.
\begin{equation}\label{LanguageModel}
P(w_1^T) = \prod\limits_{t=1}^{T} P(w_t|w_1^{t-1})
\end{equation}

The Feed-Forward Neural Network Language model (FNNLM) accomplishes the same task. Still, instead of considering all the history words, it adopts the n-gram-based idea and considers only $n-1$ words before the current word. In FNNLM, 

\begin{equation}\label{FNNLM}
\prod\limits_{t=1}^{T} P(w_t|w_1^{t-1}) \approx \prod\limits_{t=1}^{T} P(w_t|w_{t-n+1}^{t-1})
\end{equation}

Recurrent Neural Network Language Models (RNNLM) are a little different, as they have an internal state space. This internal state space can work as a memory, which stores information related to all the sequences in history and gets passed to the next sequence, enabling the model to deal with the uncertain length of sequences. 

The main concern with these language models was that they used 1-of-V encoding, also called one-hot encoding (OHE). In OHE, every word has a vector representation of size V (vocabulary size). The vocabulary size grows very fast and can reach millions of words. Representing every word with a dimension so large makes the model slow and inefficient. Furthermore, the words not seen in the training set can not be represented. These deficiencies led to dense word representation and models like \cite{collobert2008unified} and \cite{mikolov2013efficient}. A distributed representation, also termed word embedding is a real-valued representation of a word, with a lower dimension than size $V$. Each dimension in this embedding embodies a latent characteristic of the word \cite{turian2010word}. In subsequent sub-sections, we will introduce the most widespread models used to obtain these word embeddings. 
\begin{figure*}[!ht]
    \centering
    \includegraphics[width=13cm,keepaspectratio]{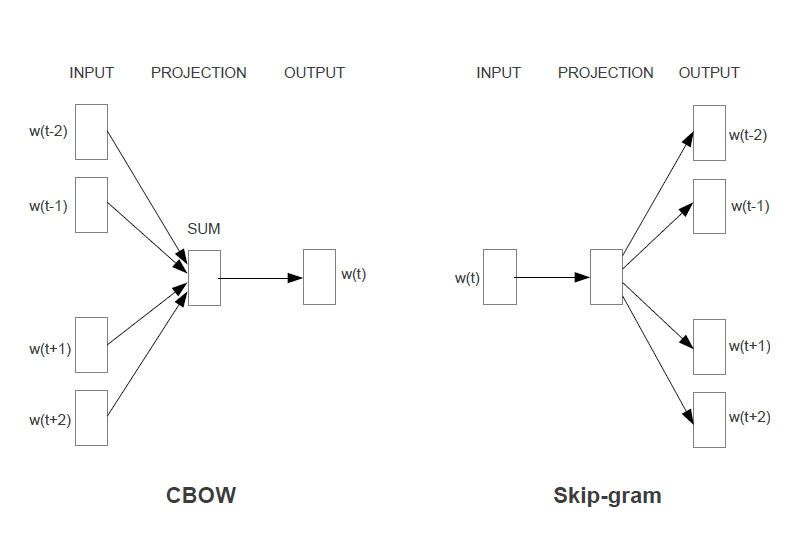}
    \caption{CBOW predicts the center word using the context words, while skipgram predicts the context words using the center word - Figure from the original word2vec paper by Mikolov}
    \label{fig:my_label}
\end{figure*}
\subsubsection{Word2vec}

NNLM got complicated over time, and later it was found that simple shallow neural network models like word2vec can work better than these overly-complicated language models. Mikolov introduced two variations of window-based neural models in 2013 - Continuous Bag Of Words (CBOW) and skipgram\cite{mikolov2013efficient, mikolov2013distributed}. 

CBOW predicts the center word using its context, so with every iteration, it tries to maximize the following probability for every word in the corpus.
\begin{equation}\label{cbow}
\sum\limits_{-c\leq{j}\leq{c},j\ne0}\log{p(w_{i}|w_{i+j})}
\end{equation}
Here, $w_i$ is the center word, while $w_{i+j}$ is the context word present at a distance $j$ from the center word. 
On the other hand, the skipgram model predicts the context words using a center word. It maximizes the following probability with each iteration. 
\begin{equation}\label{skipgram}
\sum\limits_{-c\leq{j}\leq{c},j\ne0}\log{p(w_{i+j}|w_{i})}
\end{equation}
To calculate the probability function between words, both these models use the softmax function. 

\begin{equation}\label{softmax}
p(w_c|w_o) = \frac{\exp{({v_{w_c}^{'T}  v_{w_o}})}}{\sum\limits_{w=1}^W exp({v_w^{'T} v_{w_o}})}
\end{equation}

Here \(v_w\) and $v_w^{'}$ are \textit{input} and \textit{output} vector representations of the word $w$. In both these models, we have two distinct representations for each word. For example, if we are using the skip-gram model, the input representation will be used when the word is a center word, and we will use the output representation when the word is a context word. In the end, we can either concatenate both input and output representations or take their average to construct the final word embedding. 

Mikolov introduced a few tricks in his second paper to make the model better. First, the softmax function is inefficient, as it takes sum over all the words in the vocabulary. To do this, the model becomes extensively slow for each token in our corpus. Negative sampling solves the above issue. Instead of going the whole vocabulary, we maximize the center word's similarity with the words in the context window (positive examples) and randomly select $k$ other words from the vocabulary and minimize the similarity between the center word and these $k$ negative examples. 
\begin{equation}\label{negative_sampling}
\begin{split}
p(w_c|w_o) = \log{\sigma{(v_{w_c}^{'T} v_{w_o})}} + \sum\limits_{i=1}^k{\bigg[\log{\sigma{(-v_{w_i}^{'T} v_{w_o}})}\bigg]}
\end{split}
\end{equation}

The negative samples are drawn from power distribution, and the word2vec paper suggests that the value of k can be 5-20 and 2-5 for small and large training datasets.  The second improvement was in the form of subsampling of frequent words. The authors of word2vec also note that some words (e.g., "in", "the" and "or") occur more frequently than other words while providing less or no information. The authors used a subsampling technique to tackle the issue where each word $w_i$ was discarded with some probability $P(w_i)$.
\begin{equation}\label{subsampling}
\begin{split}
P(w_i) = 1 - \sqrt{\frac{t}{f(w_i)}}
\end{split}
\end{equation}
Here $f(w_i)$ is the frequency of word $w_i$ in the corpus, and $t$ is a chosen threshold. 

While CBOW did not work very well with semantic tasks, skipgram handled semantic and syntactic tasks well. In addition to these tasks, word2vec introduced a new task of word analogies. This task examined if the produced word embeddings are able to retain the relationship between words. For example, prince is to princess as king is to queen should be retained in word embeddings as $vec(prince) - vec(princess) = vec(king) - vec(queen)$.

These models produced state-of-the-art performance in numerous tasks with a single projection layer and a simple architecture. The problem with this model was that, as it was a window-based model, it failed to use the global statistics of the corpora to a great extent. Furthermore, the word representations were not context-dependent (there was only one vector for different word senses). Global Vectors (GloVe), introduced in 2014, solves the first problem.

\subsubsection{Global Vectors (GloVe)}

GloVe\cite{pennington2014glove}, in many ways, resembles the traditional models. Like LSA, it creates the co-occurrence matrix from the text corpus but with two main differences. First, it creates a term-term co-occurrence matrix, as opposed to the term-document co-occurrence matrix in LSA, and second, instead of considering the co-occurrence count in the document, it considers the co-occurrence within a specific range or a window.  Thus, it benefits from both the corpus's global statistics and the window's local information. This property of GloVe makes it a suitable model for both word similarity and word analogy tasks. 

GloVe also introduces a new least square loss function.
\begin{equation}\label{glove_loss}
\begin{split}
J = \sum\limits_{p,q=1}^V{f(X_{pq})(w_p^T\Tilde{w}_q + b_p + \Tilde{b}_q - \log{X_{pq}})^2}
\end{split}
\end{equation}
Here $V$ is the vocabulary size, $w_p$ is the vector for center or center word, $\Tilde{w}_q$ is the vector for context word, $b_p$ is the target word bias, $\Tilde{b}_q$ is the context word bias, $X_{pq}$ is the number of times $w_p$ occurs with $w_q$ and $f(x)$ is a weighting function. The definition for the weighting function is:
\begin{equation}\label{glove_loss_second}
\begin{split}
f(x) = 
\left\{
	\begin{array}{ll}
		(x/x_{max})^\alpha  & \mbox{if } x < x_{max} \\
		1 & \mbox{otherwise}
	\end{array}
\right.
\end{split}
\end{equation}
$\alpha$ and $x_{max}$ are hyperparameters, fixed to  $0.75$ and $100$ respectively in the paper. This specific choice of weighting function has specific desiring properties such as:
\renewcommand{\labelenumi}{\alph{enumi}}
\begin{enumerate}
    \item $f(0) = 0$, as the co-occurrence becomes large, it becomes sparse and most entities become zero. Hence, $f$ should be continuous and $f(0)$ should be defined, $f(0) = 0$.
    \item It should not overweight large co-occurrences and should be relatively small for large numbers. 
\end{enumerate}

Choosing a proper loss function and utilizing both global and local information made GloVe a more suitable algorithm to model rare and frequent words. Nevertheless, the algorithm was unable to produce embeddings of unseen words, and the context of the word was still not considered. The word book had a single-word representation as opposed to having different word representations for the noun book and the verb book. Fasttext\cite{bojanowski2017enriching}, introduced in 2017, was able to solve the first issue. 

\subsubsection{Fasttext}
To model rare and previously unseen words effectively can be challenging. Besides, having a separate vector representation for each verb form can be inefficient, especially for morphologically rich languages. Fasttext solved these issues by improving the basic skipgram model and incorporating sub-word information. In fasttext, each word is represented as a sum of its character n-grams, taking into account the words' morphology. While other models such as \cite{soricut2015unsupervised} also learned morphological regularities of the language, they did not use the subword information to a large extent. An extension of the fasttext model \cite{piktus2019misspelling} also solves the problem of misspelled words, as it replaces the misspelled word with the nearest correct word. However, for our comparisons in this paper, we will take into consideration the base fasttext model. 

Their objective functions are similar as fasttext improves on the skipgram model. Fasttext starts with the skipgram objective function and improves by including the subword information. The skipgram with negative sampling objective function is, 
\begin{equation}\label{skipgram_fasttext}
\begin{split}
    \sum\limits_{t=1}^T{\bigg[ \,
    \sum\limits_{c \in C_t}{\ell(v_{w_t}^T v^{'}_{w_c})} + 
    \sum\limits_{i=1}^k{\ell{(-v_{w_t}^T v_{w_i}^{'})}
    }
    \bigg] \,}
\end{split}
\end{equation}
$v_w$ and $v_w^{'}$ are the same \textit{input} and \textit{output} word vectors while $\ell(x)$ is the logistic loss function. 

Fasttext introduces the subword information into the function, and instead of using the target word vector in the equation, it uses its n-grams' representations. Consider the word "jargon" and $n=3$, where $n$ is the n-gram length; we will have the following n-grams:\\ 
\centerline{$<$ja, jar, arg, rgo, gon, on$>$}
and the special sequence\\
\centerline{$<$jargon$>$}

The word representation is derived by summing the representations of all of its character n-grams. For example, if $\mathcal{M} \subset \{1,....,M\}$ is a word's character n-grams set and $z_m$ is the vector representation for n-gram $m$ then,
\begin{equation}\label{word_ngrams}
    \begin{split}
        v_w = \sum\limits_{m \in \mathcal{M}_w} {z_m}
    \end{split}
\end{equation}
Similarly, now the objective function of fasttext becomes,
\begin{equation}\label{fasttext}
\begin{split}
    \sum\limits_{t=1}^T{\bigg[ \,
    \sum\limits_{c \in C_t}{\ell\bigg(
    \sum\limits_{m \in \mathcal{M}_w}{(z_m^T v^{'}_{w_c})}
    \bigg)} + 
    \sum\limits_{i=1}^k{\ell{\bigg(-
    \sum\limits_{m \in \mathcal{M}_w}{(z_{m}^T v_{w_i}^{'})}
    \bigg)}
    }
    \bigg] \,}
\end{split}
\end{equation}
In simple words, this function maximizes the similarity between the n-gram representations of the center word with the context word representations, while minimizing the similarity between the n-gram representations of the center word with $k$ negative samples, for each word $t$ in the corpus $T$.  

Fasttext improved significantly on syntactic tasks, especially for morphologically rich languages like Italian, while the performance for semantic tasks remained the same. Neverthless, representing words using their n-grams helped better represent rare words and words that were not seen during training, but these representations were still context-independent. 

\subsubsection{Embeddings from Language Models(ELMo)}
In natural languages, the linguistic context of a word defines the word's meaning. The same word may have different meanings when used in different contexts, called polysemy. Methods, such as \cite{neelakantan2015efficient, huang2012improving}, have previously proposed models to overcome the problem of polysemy in learning word representations by learning multiple embeddings for each word. Nevertheless, these models had shallow architectures and required predefined word sense classes. ELMo is a deep contextualized model for learning word embeddings from unlabeled text. ELMo representations are deep and contextualized, the reason being that they are a function of a bidirectional model's internal layers, and they depend on the context of the word. ELMo word embeddings are a function of the entire sentence. 

A forward language model takes the tokens $(t_1,...,t_{i-1})$ and models the probability of $t_i$ to predict the likelihood of the token sequence,
\begin{equation}\label{forwardlanguagemodel}
    \begin{split}
        p(t_1,t_2,...,t_N) = \prod\limits_{i=1}^N {p(t_i | t_1,...,t_{i-1})}
    \end{split}
\end{equation}
The model takes a context-independent token embedding and run in through $L$ layers of forward Long Short Term Memory networks (LSTMs). Every layer $j$ outputs a context dependent representation of the token $t_i$, $\overrightarrow{h}_{i,j}^{LM}$\cite{jozefowicz2016exploring, melis2017state}. The LSTM output for the last layer can be used for predicting the next token $t_{i+1}$. A backward language model, on the other hand, uses the future tokens $(t_{i+1}, t_{i+2},...,t_N)$ to model the likelihood of the current token $(t_i)$
\begin{equation}\label{backwardlanguagemodel}
    \begin{split}
        p(t_1,t_2,...,t_N) = \prod\limits_{i=1}^N {p(t_i | t_{i+1}, t_{i+2},...,t_N)}
    \end{split}
\end{equation}
The same way as above, $L$ layers of backward LSTMs is implemented, and each layer $j$ computes a context dependent hidden representation of the token $t_i$, $\overleftarrow{h}_{i,j}^{LM}$.

A biLM couples a forward language model and a backward language model, and ELMo simultaneously maximizes the log-likelihood of the biLM:
\begin{equation}\label{biLM}
    \begin{split} 
{\sum\limits_{i=1}^N{(\log{p(t_i |t_1,...,t_{i-1}; \Theta_{x}, \overrightarrow{\Theta}_{LSTM},\Theta_{s})}}}\\
+ \log {p(t_i | t_1,...,t_{i-1};\Theta_{x},\overleftarrow{\Theta}_{LSTM},\Theta_{s}))}
    \end{split}
\end{equation}
$\Theta_x$ and $\Theta_s$ are parameters for token representation and the softmax layer. $\overrightarrow{\Theta}_{LSTM}$ and $\overleftarrow{\Theta}_{LSTM}$ are the parameters for forward and backward LSTMs. 

The model computes $2L+1$ parameters, a token representation $x_i$, and a hidden representation $h_{i,j}^LM$ for each layer of the forward and backward LSTMs, for each token $t_i$. If $x_{i}^{LM} = h_{i,0}^{LM}$ and $h_{i,j}^{LM} = [\overrightarrow{h}_{i,j}^{LM}; \overleftarrow{h}_{i,j}^{LM}]$, for each biLSTM layer, then 
\begin{equation}\label{elmoparameters}
    \begin{split}
        R_i = \{
         h_{i,j}^{LM} | j = 0,...,L
        \}
    \end{split}
\end{equation}
In the simplest case, ELMo will select the top layer, $E(R_i) = h_{i,L}^{LM}$, as in \cite{peters2017semi}, and more generally, it will estimate a weighting of all biLM layers for the job in hand.

ELMo addressed many issues present in the understanding of natural language, including polysemy. It had a deep architecture as opposed to the previous word representation models like word2vec and GloVe, and the produced representations were context-aware. After ELMo, many similar deep pre-trained models, for example: Open AI GPT\cite{radford2018improving}, Open AI GPT2\cite{radford2019language}, and BERT\cite{devlin2018bert}, were introduced.

\begin{figure*}[!ht]
    \centering
    \includegraphics[width=13cm,keepaspectratio]{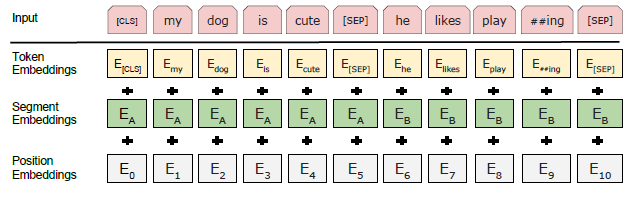}
    \caption{The BERT input representations are obtained by summing the token embeddings, position embeddings, and the segmentation embeddings - Figure from BERT original paper by Devlin et al.,}
    \label{bert_encoding}
\end{figure*}
\begin{center}
    \begin{figure*}[!ht]
        \includegraphics[width=13cm, keepaspectratio]{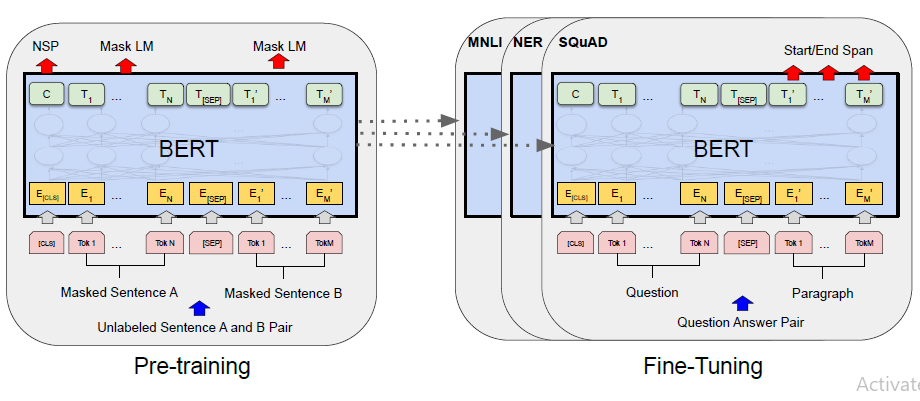}
        \caption{The BERT training phases. Apart from output layers, both pre-training and fine-tuning use the same architecture. During fine-tuning, the model is initialized with pre-trained model parameters and then fine-tuned for task in hand. Figure from the BERT original paper. }
        \label{bert_phases}
    \end{figure*}
\end{center}

\subsubsection{Bidirectional Encoder Representations from Transformers (BERT)}
BERT is a multi-layer bidirectional Transformer encoder \cite{vaswani2017attention} based language representation model. It uses a Masked Language Model (MLM) to encode both left and right contexts into token representation. MLM masks some of the input tokens randomly and then predict the vocabulary ID for masked tokens. Implementing BERT involves two steps: pre-training and fine-tuning. In the pretraining phase, the model is trained on a large corpus of unlabeled text. During the fine-tuning phase, the parameters learned in the pretraining phase are initialized and subsequently fine-tuned for a particular downstream task. 

BERT handles various downstream tasks due to its flexible architecture. It can represent a single or two sentences paired together $\big(Sentence A, Sentence B\big)$. To represent two sentences, BERT adds a unique token, $[SEP]$. Additionally, every sequence starts with a particular classification token $[CLS]$. We use the final state of $[CLS]$ token as a sequence representation for classification tasks. A token's input representation is obtained by summing the corresponding token embedding, segmentation embedding, and positional embedding. For token embeddings, BERT uses WordPiece embedding \cite{wu2016google}, segmentation embedding of the token denotes whether a token belongs to sentence A or sentence B, and positional embedding contains the token's position in the input sequence. BERT proved very useful in language modeling and outperformed other models in almost every language modeling task. 

\begin{center}
    \begin{table*}[!t]
         \begin{tabularx}{\linewidth}{m{1.6cm}|X|X|X|X|X|X}
         \hline
              Approaches & Density & Polysemy & Context-awareness & Fine-tuning & OOV word vectors & Resource intensity  \\
         \hline 
              One-hot & \centering{$\times$} & \centering{$\times$} & \centering{$\times$} & \centering{$\times$} & \centering{$\times$} & \Centering3 
              \\
              LSA & \centering{\checked} & \centering{$\times$} & \centering{$\times$} & \centering{$\times$} & \centering{$\times$} & \Centering3
              \\
              Word2vec & \centering{\checked} & \centering{$\times$} & \centering{$\times$} & \centering{$\times$} & \centering{$\times$} & \Centering1
              \\
              GloVe & \centering{\checked} & \centering{$\times$} & \centering{$\times$} & \centering{$\times$} & \centering{$\times$} & \Centering1
              \\
              Fasttext & \centering{\checked} & \centering{$\times$} & \centering{$\times$} & \centering{$\times$} & \centering{\checked} & \Centering2
              \\
              ELMo & \centering{\checked} & \centering{\checked} & \centering{\checked} & \centering{\checked} & \centering{\checked} & \Centering4
              \\
              BERT & \centering{\checked} & \centering{\checked} & \centering{\checked} & \centering{\checked} & \centering{\checked} & \Centering5
              \\
              \hline
         \end{tabularx}
         \caption{The Table shows the presence and absence of certain properties in different word embedding approaches. The resource intensity is lower for the models that require lesser memory and training time.}
        \label{tab:1}
    \end{table*}
\end{center}
\section{Properties of Existing Word Embedding Approaches}

We start evaluating the existing word representation approaches by comparing their intrinsic properties before external evaluations. These properties affect the performance of these approaches on extrinsic tasks directly or indirectly. We start from the basic word representation approach (e.g., One-hot encoding) and take into consideration the existing state-of-the-art word embedding models. We study these models concerning properties such as the density of the word representations, polysemy, context-awareness, fine-tuning, resource intensity, and representing out of vocabulary (OOV) words. 
 
The density of the word representations, generated by a word embedding model, makes the model memory efficient. It also enables the model to encode latent features corresponding to the words instead of encoding plain statistical information. While initial models (e.g., one-hot encoding) generates sparse representations, other models (e.g., LSA) generates sparse representations and then projects them to lower dimensions. Modern approaches, on the other hand, starts with learning dense low dimensional word vectors. Generating low-dimensional word vectors enables the word embedding models to encode hidden properties of the words, hence preserving the semantic and syntactic regularities present in the corpus. For example, word2vec maximizes log probability between words that lie closer to each other. hence, words with the same context tend to have \textit{similar} or \textit{closer} word representations. 

Polysemy enables the word embedding models to learn different word vectors for different word senses or even the same sense. A word may provide different meanings when used in different contexts. A word embedding model should be able to keep track of word context to generate word vectors for polysemous words. Traditional models (e.g., LSA and one-hot encoding) provide a single static word representation for each word, irrespective of its sense. GloVe, word2vec, and fasttext also provide a similar static word representation, not considering the polysemy. ELMo and BERT, on the other hand, provide distinct word representations for each word sense. This is possible, as these models keep track of the context of the word. Context-dependent word representations further improve the performance of ELMo and BERT on downstream tasks, as the meaning of the word affects its representations. Consider a word \textit{book}. It may be used as a \textit{noun} or a \textit{verb}. In the former case, the representation of the word should be closer to words like \textit{notebook} and \textit{pen}, while in the latter case, the representation of the word should be closer to other words such as \textit{ticket}. Hence, multiple context-dependent representations enable models like BERT and ELMo to encode and effectively leverage certain words' meanings. 

Fine-tuning allows the model to tune its parameters for the task. Word embedding models generate word representations, which can be used as features for underlying classification tasks. While some models (e.g., GloVe and word2vec) generate fixed word vectors, others (e.g., BERT and ELMo) allow fine-tuning to generate more appropriate word representations for the task.  

OOV words are those words that the model hasn't seen during training. The model can produce a word representation for an unseen word if it understands the morphology of the words in some manner. While traditional models, word2vec and GloVe, do not utilize the sub-word information and act on each word as a single indivisible entity, modern approaches such as fasttext, ELMo, and BERT can benefit from the sub-word information. Fasttext perceives every word as a combination of n-grams, which enables fasttext to represent OOV words. The n-grams required to build the word may previously exist, and the representation of the word is the sum of all its n-grams. ELMo can handle OOV word representation as it uses character embeddings to computer word-level embeddings. BERT uses the WordPiece algorithm to divide a word into \textit{pieces} and represent each \textit{piece} with an embedding.  

While most of the focus goes on how different word embedding approaches perform in various tasks, it is necessary to know the resource intensity of these approaches. Sometimes, a model consumes much more time and resources compared to a different model, while their performance on a task may not differ significantly. The comparison of different word representation models concerning their intrinsic properties is presented in Table \ref{tab:1}.  

\section{Experimental Results}
\subsection{Datasets}
We evaluate the models mentioned above concerning 4 different classification tasks: spam detection, radical language detection, abusive language detection, and distinguishing abusive and hateful language. The spam detection and radical language detection datasets both contain 200K (100K negative tweets and 100K positive tweets) each, the abusive language detection dataset contains 25K tweets (12.5K negative tweets and 12.5K positive tweets), and the abusive language vs. hateful language dataset contains 5K tweets (2.5K abusive language tweets and 2.5K hateful language tweets). The datasets details have been given in table \ref{table:2}. 

We have conducted different experiments in multiple settings for the spam detection and radical language detection datasets. For example, we have evaluated the models with different subsets of these datasets (25K, 50K, 100K, 150K, and 200K tweets). We have also compared the models using different ratios of the positive and negative classes to examine the impact of data skewness on each model. 
\begin{table}[h!]
\renewcommand{\arraystretch}{1.3}
\begin{tabular}{m{4cm}m{7.4cm}}
    \hline
     Dataset & Explanation \\
     \hline
     Spam-N & A subset of spam detection dataset with N tweets. It may be balanced (1:1), or unbalanced (n:1), containing n negative normal tweets for every positive spam tweet. \\
     \hline
     Radical-N & A subset of radical language dataset with N tweets. It may be balanced (1:1), or unbalanced (n:1), containing n negative nromal tweets for every positive radical tweet. \\
     \hline
     Abusive language & This dataset contains 25K tweets. The dataset is balanced, equally distributed between positive and negative tweets.\\
     \hline
     Abusive vs. hateful &This Dataset contains 5K tweets. The dataset is balanced, containing the same number of abusive language and hateful language tweets. \\
     \hline
\end{tabular}
\caption{A brief description of each dataset. }
\label{table:2}
\end{table}

\subsection{Experimental model setup}
Our experimental model consists of an input embedding layer, two bi-LSTM layers of 32 and 64 units, respectively, a 128-unit dense layer, and a sigmoid output layer. 

We trained our models using the tensorflow framework \cite{abadi2016tensorflow}. We used the gensim library \cite{rehurek_lrec} to train word2vec and fasttext vectors on our corpora. We also used a Tensorflow implementation of GloVe to train our vectors instead of using the original code written in C. For ELMO and BERT, we used tensorflow-hub to retrieve the context-dependent pre-trained vectors. In our experiments, we used the BERT-base (L=12, H=768) implementation of the BERT model.

\subsection{Evaluating Trained Word Embeddings }
While training word vectors on our corpora, there are many hyper-parameters such as window size and vector dimension. Choosing the correct value for a hyper-parameter can drastically affect the computed word vectors. We have trained word vectors for word2vec (skipgram and CBOW), GloVe, and fasttext on our corpora. Afterward, we have used these computed word vectors for our classification tasks and report the results. We have not trained vectors for BERT and ELMo from scratch, as training these vectors from scratch is inefficient compared to pre-trained vectors considering the smaller corpus size. Additionally, they are resource and time-exhaustive, and as we will see in the next section, it is better to use their pre-trained vectors.  

\subsubsection{Window Size: } Window size refers to the number of words to both sides of the center word that we consider when we train our word embeddings. In skipgram, CBOW, and fasttext, the window size is used to maximize the log probability between the center and the context words (words present in the window), while in GloVe, we use the window size to construct the word co-occurrence matrix. We have selected three different window sizes and reported the result for each dataset in table \ref{table:3}.

While GloVe performs better with large window sizes consistently, other models tend to have an up and down. Although the difference between the performance on different window sizes for all the models is not significant, we can safely say that the models can capture more information with large window sizes. We should keep in mind that large window sizes require more computational and memory resources. Also, with much larger window sizes, the model may capture irrelevant information. For example, a word may not depend or be relevant to another word that occurred 20 words later.

\begin{landscape}
\begin{center}
    \begin{table*}[!ht]
    \begin{tabularx}{\linewidth}{|m{3cm}|c|c|c|c|c|c|c|c|c|c|c|c|}
        \hline
          & \multicolumn{3}{|c|}{Skipgram}& \multicolumn{3}{|c|}{CBOW} & \multicolumn{3}{|c|}{GloVe} &\multicolumn{3}{|c|}{fasttext}\\ 
         \hline
         Datasets & w=2 & w=5 & w=10 & w=2 & w=5 & w=10 & w=2 & w=5 & w=10 & w=2 & w=5 & w=10 \\
         \hline
         Spam-100K & 0.823 & 0.820 & \textbf{0.828} &
         0.828 & 0.84 & \textbf{0.841} &
         0.849 & 0.848 & \textbf{0.85} &
         \textbf{0.854} & 0.844 & 0.849
         \\
         \hline
         Spam-100K (10:1)& 0.510 & 0.517 & \textbf{0.529} &
         \textbf{0.57} & 0.56 & 0.569 &
         0.597 & 0.597 & \textbf{0.60} &
         \textbf{0.593} & 0.567 & 0.56
         \\
         \hline
         Spam-30K & 0.808 & \textbf{0.821} & 0.818 &
         \textbf{0.814} & 0.804 & 0.798 &
         0.82 & 0.82 & \textbf{0.823} &
         0.809 & \textbf{0.818} & 0.80
         \\
         \hline
         Radical-100K & \textbf{0.999} & 0.998 & 0.998 &
         \textbf{0.999} & 0.998 & 0.998 &
         0.995 & 0.995 & \textbf{0.997} &
         0.997 & 0.997 & \textbf{0.998} 
         \\
         \hline
         Radical-30K & 0.997 & 0.997 & 0.997 &
         \textbf{0.998} & 0.996 & 0.996 &
         0,994 & 0.995 & \textbf{0.996} &
         0.997 & 0.996 & \textbf{0.998}  
         \\
         \hline
         Abusive language & 0.993 & \textbf{0.995} & 0.994 &
         \textbf{0.992} & 0.991 & 0.991 &
         0.980 & 0.982 & \textbf{0.985} &
         0.993 & 0.993 & \textbf{0.994}
         \\
         \hline
         Abusive vs. hateful & 0.733 & 0.741 & \textbf{0.767} &
         0.718 & 0.709 & \textbf{0.720} &
         0.768 & 0.767 & \textbf{0.769} &
         0.745 & \textbf{0.771} & 0.741 
         \\
         \hline
    \end{tabularx}
    \caption{The table shows the performance of different models, trained on different window sizes, on external classification tasks. All models have been trained with 300 dimension vectors on the same task's training data. We have reported f1 score for the positive class for every task. All datasets, except the ones we have mentioned the ratio with, are balanced. }
    \label{table:3}
    \end{table*}
\end{center}

\begin{center}
    \begin{table*}[!ht]
    \begin{tabularx}{\linewidth}{|m{4cm}|c|c|c|c|c|c|c|c|c|c|c|c|}
        \hline
         - & \multicolumn{3}{|c|}{Skipgram}& \multicolumn{3}{|c|}{CBOW} & \multicolumn{3}{|c|}{GloVe} &\multicolumn{3}{|c|}{fasttext}
          \\ 
         \hline
         Datasets & 50d & 200d & 300d & 50d & 200d & 300d & 50d & 200d & 300d & 50d & 200d & 300d 
         \\
         \hline
         Spam-200K & 0.853 & 0.859 & 0.856 &
         0.859 & 0.860 & 0.862 &
         0.821 & 0.855 & 0.86 &
         0.863 & 0.866 & 0.866
         \\
         \hline
         Spam-100K & 0.833 & 0.835 & 0.820 &
         0.84 & 0.84 & 0.84 &
         0.843 & 0.847 & 0.848 &
         0.854 & 0.856 & 0.844 
         \\
         \hline
         Spam-30K & 0.805 & 0.80 & 0.824 &
         0.811 & 0.81 & 0.804 &
         0.811 & 0.814 & 0.82 &
         0.808 & 0.818 & 0.818
         \\
         \hline
         Radical-100K& 0.997 & 0.997 & 0.998 &
         0.998 & 0.998 & 0.998 &
         0.995 & 0.994 & 0.995 &
         0.996 & 0.996 & 0.997 
         \\
         \hline
         Radical-30K & 0.995 & 0.995 & 0.997 &
         0.996 & 0.997 & 0.996 &
         0.995 & 0.994 & 0.995 &
         0.997 & 0.996 & 0.996 
         \\
         \hline
         Abusive language & 0.995 & 0.995 & 0.995 &
         0.990 & 0.990 & 0.991 &
         0.978 & 0.979 & 0.982 &
         0.991 & 0.992 & 0.993 
         \\
         \hline
         Abusive vs. hateful & 0.750 & 0735 & 0.741 &
         0.721 & 0.739 & 0.709 &
         0.722 & 0.755 & 0.767 &
         0.763 & 0.78 & 0.771 
         \\
         \hline
        \end{tabularx}
        \caption{The table shows the effect of trained word vector dimension on extrinsic classification tasks. All models have been trained with a window size of 5.  }
    \label{table:4}
    \end{table*}
\end{center}

\end{landscape}

\subsubsection{Embedding Dimension} 
We have also trained word vectors of various dimensions, to examine its effect on intrinsic classification tasks. While for most of these tasks, the performance is similar for high dimension and low dimension vectors, we witness that for tasks with more data (e.g., Spam-200K), the higher dimension vectors perform better. High-dimension vectors can encode more data and represent words more sparsely, resulting in better performance for most extrinsic tasks. On the contrary, if the data is less, smaller dimension vectors can encode the information better in some scenarios (e.g., Abusive vs. hateful). 

Ultimately, higher dimension vectors may perform similarly, if not better, for almost all tasks, but they also require more memory and time to train. The F1 scores for multiple datasets have been presented in table \ref{table:4}.
\begin{center}
    \begin{table*}[!b]
        \begin{tabularx}{\linewidth}{|m{3cm}|X|X|X|X|X|}
        \hline
          Datasets & 
          \multicolumn{1}{|c|}{Word2vec}& 
          \multicolumn{1}{|c|}{GloVe} &
          \multicolumn{1}{|c|}{fasttext} & 
          \multicolumn{1}{|c|}{ELMo}
          & \multicolumn{1}{|c|}{BERT}
          \\ 
          \hline
         Spam-100K & 0.822 & 0.818 & 0.836 & 0.969 &  0.979
         \\
         \hline
         Radical-100K & 0.995 & 0.995 & 0.998 & 0.999 &  1.0
         \\
         \hline
         Abusive language & 0.982 & 0.982 & 0.993 & 0.997 & 1.0
         \\
         \hline
         Abusive vs. hateful & 0.771 & 0.782 & 0.783 & 0.761 & 0.833
         \\
         \hline
        \end{tabularx}
        \caption{This table shows the performance of all the various pre-trained word embedding models on various classification tasks. }
    \label{table:5}
    \end{table*}
\end{center}

\subsection{Evaluating Pre-trained Word Embeddings}
We have also used the pre-trained vectors for extrinsic classification tasks and reported the results. In addition to word2vec, GloVe, and fasttext, we have included BERT and ELMo in the evaluation in this section. Pre-trained word vectors for word2vec, GloVe, and fasttext are static, as they are context-independent. At the same time, BERT and ELMo provide context-dependent word vectors and shall be extracted for each sentence as we train our classification model. We will first compare these models' performance on certain classification tasks and subsequently examine the effect of some factors on these models. For our experiment, we have used BERT-base, which has 768 dimension vectors, ELMo 1024 dimension vectors, and all other models have 300 dimensions word representations.

Table 5 shows the F1 score on various classification tasks for all the models. We can see that all models perform very well on relatively easy datasets (e.g., radical language and abusive language), while many models struggle on others. Differentiating abusive tweets from hate tweets is an arduous task, as both contain similar language structure, even then BERT improves the F! score for the dataset by almost 5\% compared to any other model.

\subsubsection{Dataset Size } In this segment, we evaluate the effect of the data size on pre-trained word embeddings' performance. We select multiple subsets of the datasets and evaluate the models, examining their classification F1 score. While it may be evident that more data will result in better performance, there are situations where a massive amount of data may not be available. It is also essential to know the magnitude of the effect of classification data increment on the model performance.  

All models have been pre-trained with billions of words. While BERT and ELMo possess task-specific fine-tuning abilities, other models do not. The effect of these two features, fine-tuning and context-dependency, is noticeable, as these two models perform extraordinary well on the extrinsic classification tasks. Furthermore, while BERT barely suffers from smaller classification datasets, the loss for other models is relatively high. Word2vec suffers the most, as its F1 score decreases by more than 6 percent for spam classification when the dataset contains 25K instead of 200K tweets. 

The classification results further prove that both datasets have a very different text structure. While Radical tweets are well structured and more formally written, all models perform extraordinarily well while classifying them. Furthermore, radical tweets contain specific words that can rarely be found in regular tweets, which further differentiates them from regular tweets' structure. On the other hand, spam tweets are more similar to normal tweets, and most of these spam tweets contain slang language and abbreviations for which there are no representations in many models. 

Another point that further helps BERT is the mechanism in which BERT divides a word into multiple pieces using the WordPiece algorithm at its heart. This partition helps BERT provide context-dependent vectors for those words, which may otherwise be by OOV words. Table \ref{table:6} depicts the results.
\begin{center}
    \begin{table*}[!b]
        \begin{tabularx}{\linewidth}{|m{2.1cm}|c|c|c|c|c|c|c|c|}
        \hline
          Datasets & \multicolumn{4}{|c|}{Spam}& \multicolumn{4}{|c|}{Radical}
          \\ 
          \hline
            & 50K & 100K & 150K & 200K &
            50K & 100K & 150K & 200K
         \\
         \hline
         Word2vec  & 0.81 & 0.82 & 0.83 & 0.85 &
             0.99 & 0.99 & 0.99 & 1.0
         \\
         \hline
         GloVe & 0.81 & 0.82 & 0.83 & 0.84 &
          0.99 & 0.99 & 0.99 & 1.0
         \\
         \hline
         fasttext & 0.81 & 0.84 & 0.84 & 0.85 &
          1.0 & 1.0 & 1.0 & 1.0
         \\
         \hline
         ELMo & 0.96 & 0.97 & 0.97 & 0.98 &
          1.0 & 1.0 & 1.0 & 1.0
         \\
         \hline
         BERT & 0.98 & 0.98 & 0.98 & 0.98 &
         1.0 & 1.0 & 1.0 & 1.0
         \\
         \hline
        \end{tabularx}
        \caption{This table shows the effect of an increase or decrease in the dataset size on the performance of word embedding models. BERT vectors have 768 dimensions, ELMo uses 1024 dimension vectors, and all other models have 300 dimension pre-trained vectors. }
    \label{table:6}
    \end{table*}
\end{center}

\subsubsection{Unbalanced Data } Deep learning models usually tend to overfit towards the more general class. This section compares all word embedding models based on their performance on unbalanced data. Although all models perform worse on skewed data, BERT and ELMo keep their performance above par on the most complex tasks. Once again, we observe that with the easier dataset, the influence of data skewness is negligible, while with the complex task (e.g., spam classification), the impact is more prominent. The classification F1 score for various datasets is presented in table \ref{table:7}.

    \begin{table*}[!ht]
        \begin{tabularx}{\linewidth}{|m{3cm}|X|X|X|X|X|X|X|X|}
        \hline
          Datasets & \multicolumn{4}{|c|}{Spam}& \multicolumn{4}{|c|}{Radical}
          \\ 
          \hline
           & 1:1 & 2:1 & 5:1 & 10:1 & 
           1:1 & 2:1 & 5:1 & 10:1
         \\
         \hline
         Word2vec & 0.822 & 0.759 & 0.646 & 0.539 &
         0.995 & 0.994 & 0.990 & 0.986
         \\
         \hline
         GloVe & 0.818 & 0.763 & 0.647 & 0.543 &
         0.995 & 0.994 & 0.992 & 0.990 
         \\
         \hline
         fasttext & 0.836 & 0.771 & 0.666 & 0.613 &
         0.998 & 0.997 & 0.996 & 0.996 
         \\
         \hline
         ELMo & 0.969 & 0946 & 0.945 & 0.859 &
         0.999 & 0.998 & 0.997 & 0.994
         \\
         \hline
         BERT & 0.979 & 0.965 & 0.942 & 0.922 &
         1.0 & 1.0 & 1.0 & 1.0 
         \\
         \hline
        \end{tabularx}
        \caption{F1 score on balanced and unbalanced datasets. }
    \label{table:7}
    \end{table*}
\subsubsection{Classification Data Pre-processing } 
While pre-processing is an integral part of training deep learning classification models, we would like to examine the effect of fully and partially cleaning the classification data on classification results. 

In the first case, we clean the data completely. We remove URLs, hashtags, symbols, mentions, and stopwords from the tweets. We also lemmatize the words. While in the latter case, we only remove URLs and hashtags from the data and do not lemmatize the words.

\begin{center}
    \begin{table}[ht]
    \begin{tabularx}{\linewidth}{|X|X|X|X|X|}
         \hline
         Datasets & \multicolumn{2}{|c|}{Spam-100K}
         & \multicolumn{2}{|c|}{Radical-100K}
         \\
         \hline
          & F-Cl & P-Cl
          & F-Cl & P-Cl
          \\
          \hline
          word2vec & 0.791 & 0.822
           & 0.988 & 0.995
          \\
          \hline
          GloVe & 0.792 & 0.818
          & 0.989 & 0.995 
          \\
          \hline
          fasttext & 0.788 & 0.836 
          & 0.988 & 0.998 
          \\
          \hline
          ELMo & 0.801 & 0.969
           & 0.993 & 0.999
          \\
          \hline
          BERT &
           0.822 & 0.979 
          & 0.995 & 1.0 
          \\
          \hline
    \end{tabularx}
    \caption{F1 score for partially and fully cleaned data while classification. F-Cl stands for fully cleaned, and P-Cl stands for partially cleaned. }
    \label{table:8}

    \end{table}
\end{center}

The partially cleaned data yields better results in all cases. This indicates the stopwords and actual verb form on the classification result. We also observe that the boost in performance for fasttext, ELMo, and BERT while partially cleaning the data, is higher. The reason behind this boost of performance is the use of sub-word information in these models, which allows them to represent OOV words and use the morphology of the words. The results are shown in table \ref{table:8}

\subsection{Multi-class Classification}
Until now, we have classified tweets into two classes: positive and negative. We need to classify our data into multiple categories or classes in some scenarios. Here, we evaluate the pre-trained word embeddings' performance in multi-class classification. 

We build a new dataset, consisting of five classes, and 2500 tweets for every class. We also make changes to our classification model, changing the sigmoid output layer to a softmax output layer with five units to accommodate multi-class classification.  The complete comparison is presented in Table \ref{table:9}.

We present the results for both fully cleaned and partially cleaned data. While with fully cleaned data, there is a slight difference between BERT and GloVe; the difference is much more noticeable with partially cleaned data. This further emphasizes our previous point that fully cleaning the data results in information loss.

\begin{center}
    \begin{table}[ht]
    \begin{tabularx}{\linewidth}{|m{3cm}|X|X|}
         \hline
         Datasets & \multicolumn{2}{|c|}{Multi-Class}
         \\
         \hline
          & F-Cl & P-Cl
          \\
          \hline
          word2vec & 0.730 & 0.779
          \\
          GloVe & 0.757 & 0.798
          \\
          fasttext & 0.714 & 0.766 
          \\
          ELMo & 0.745 & 0.856
          \\
          BERT &
           0.777 & 0.873 
          \\
          \hline
    \end{tabularx}
    \caption{Macro F1 score for Multi-Class classification. The dataset consists of 5 classes, with 2500 tweets for each class. F-Cl stands for fully cleaned, and P-Cl stands for partially cleaned. }
    \label{table:9}
    \end{table}
\end{center}

\begin{center}
    \begin{table*}[ht]
    \begin{tabularx}{\linewidth}{|m{3cm}|X|X|X|X|X|X|}
         \hline
         & \multicolumn{2}{|c|}{word2vec} &
         \multicolumn{2}{|c|}{GloVe} &
         \multicolumn{2}{|c|}{fasttext}
         \\
         \hline
         Dataset & Pre-trained & trained
         & Pre-trained & trained
         & Pre-trained & trained
         \\
         \hline
         Spam-200K & 0.849 & 0.856
         & 0.844 & 0.86
         & 0.854 & 0.866
         \\
         \hline
         Spam-100K & 0.822 & 0.82 
         & 0.818 & 0.848
         & 0.836 & 0.844
         \\
         \hline
         Spam-30K & 0.79 & 0.821
         & 0.788 & 0.82
         & 0.816 & 0.818
         \\
         \hline
         Radical-100k & 0.995 & 0.998
         & 0.995 & 0.995
         & 0.998 & 0.997
         \\
         \hline
         Radical-30K & 0.995 & 0.997
         & 0.995 & 0.995
         & 0.998 & 0.996
         \\
         \hline
         Abusive language & 0.982 & 0.995
         & 0.982 & 0.982
         & 0.993 & 0.993
         \\
         \hline
         Abusive vs. hateful & 0.771 & 0.741
         & 0.782 & 0.767
         & 0.783 & 0.771
         \\
         \hline
    \end{tabularx}
    \label{table:10}
    \caption{F1 score for pretrained and trained word embeddings on various datasets.}
    \end{table*}
\end{center}

\subsection{Pretrained Word Embeddings vs. Trained Word Embeddings}
We evaluated context-independent trained word embedding models. We also compared all major pre-trained word embedding models on multiple extrinsic classification tasks. Now, we will compare pre-trained word embeddings generated by these context-independent models with the word embeddings that we have trained on our corpora. We have used similar possible hyper-parameter values. We use 300 dimension vectors, with a context size of 5, and for word2vec, we use the skipgram model with a negative sampling value of 15. 

For most of the tasks, the word representation that we have trained from scratch outperforms the pre-trained word embeddings. This may be because the corpus used for training these word embeddings may not be relevant to the task in hand. Another reason can be that the corpus contain some words, for which the embedding is not availabe in the pre-trained embeddings. Although if we have a small classification dataset (e.g., abusive vs. hateful), it would be better to use pre-trained vectors, as small text corpus will not generate quality word representations. The data factor becomes more evident once we examine different subsets of the spam dataset. While the performance difference between the pre-trained and trained vectors is more for the smaller subset (30K tweets), it decreases considerably for the bigger subsets (100K and 200K tweets). See Table 10 for details.

\section{Conclusion}
This paper evaluated the existing word embedding algorithms on extrinsic classification tasks. We also described the working of each of these models and provided insight into how these models encode the relations between words. We explained the desired properties of a good word embedding approach and discussed the presence and absence of these properties in certain models. We also illustrated how these properties affect the word embeddings produced by word embedding models. The impact of certain parameters such as., window size, embedding dimension, was also illustrated on the word embeddings' quality. We also compared pre-trained word embeddings, trained word embeddings, and their impact on classification tasks. Moreover, we also obtained an insight into which algorithm performs better when used for multi-class classification. 

Although studying all the word embedding algorithms and their properties in a single paper is difficult, we covered the most important approaches, selected from different categories. To simplify our comparisons, we divided the models into two groups, traditional and neural models, and observed that neural word embedding models provide numerous advantages over traditional approaches. 

We observed that while BERT overperformed the other word embedding approaches in almost every task, in certain classification tasks (e.g., Abusive vs. Hateful), the difference was negligible. Hence, considering BERT's resource extensiveness, simpler models (e.g., word2vec and GloVe) provided better results. ELMo, on the other hand, performed similarly to BERT in most tasks, as it uses task-specific parameter tuning. 

We also observed that the structure of underlying classification data plays a vital role. While some datasets (e.g., Radical dataset), may be very easy to classify, as their text combinations are very different, in other datasets (e.g., Abusive vs. hateful), both the classes follow almost the same text pattern, which makes it hard to differentiate. Before selecting a word embedding model, we shall analyze the underlying classification data and examine the text structure of different text classes. When the classes follow different text patterns, the simple models (e.g., word2vec and GloVe) perform similarly to complex models (e.g., BERT). 

Data pre-processing, although an integral part of a classification task, can impact the result negatively if done extensively. We observed that the original morphology of the text provides crucial information and can be lost if the pre-processing is done carelessly. The above point further emphasizes the importance of using sub-word information. Approaches that consider words as indivisible entities can face OOV words and hence an increasing number of unknown word vectors. All these unknown word vectors result in a loss of information. Approaches that use sub-word or character-level information, on the other hand, can handle OOV words and represent the actual morphology of the words effectively. Hence models like BERT, ELMo and fasttext, perform better than their predecessors.

\printbibliography
\end{document}